\documentclass[letterpaper, 10pt, conference]{ieeeconf}  

\IEEEoverridecommandlockouts                              

\usepackage{times}
\usepackage{epsfig}
\usepackage{graphicx}
\usepackage{amsmath}
\usepackage{bm}
\usepackage{amssymb}
\usepackage{listings}
\usepackage{multirow}
\usepackage{subfig}
\usepackage[colorlinks,citecolor=blue]{hyperref}

\newcommand{\etal}{\mbox{\emph{et al.\ }}}
\newcommand{\ie}{\mbox{\emph{i.e.\ }}}
\newcommand{\eg}{\mbox{\emph{e.g.\ }}}

\title{\LARGE \bf
An Efficient Volumetric Mesh Representation\\ for Real-time Scene Reconstruction using Spatial Hashing
}

\author{Wei Dong, Jieqi Shi, Weijie Tang, Xin Wang, and Hongbin Zha
\thanks{The authors are with the Key Laboratory of Machine Perception, Peking University, China. {\tt\small \{w.dong, jaycee\_sjq, twjjwt, xinwang\_cis\}@pku.edu.cn, zha@cis.pku.edu.cn}. This research was supported by Natural Science Foundation of China (No. 61632003).}%
}

\begin{document}

\maketitle
\thispagestyle{empty}
\pagestyle{empty}

\begin{abstract}
	Mesh plays an indispensable role in dense real-time reconstruction essential in robotics. Efforts have been made to maintain flexible data structures for 3D data fusion, yet an efficient incremental framework specifically designed for online mesh storage and manipulation is missing. We propose a novel framework to compactly generate, update, and refine mesh for scene reconstruction upon a volumetric representation. Maintaining a spatial-hashed field of cubes, we distribute vertices with continuous value on discrete edges that support $O(1)$ vertex accessing and forbid memory redundancy. By introducing Hamming distance in mesh refinement, we further improve the mesh quality regarding the triangle type consistency with a low cost. Lock-based and lock-free operations were applied to avoid thread conflicts in GPU parallel computation. Experiments demonstrate that the mesh memory consumption is significantly reduced while the running speed is kept in the online reconstruction process.
\end{abstract}

\section{Introduction}
Due to the appearance of light-weight, consumer level depth sensors such as Kinect and Structure Sensor, on-the-fly dense reconstruction of ordinary scenes has become a popular topic. In the field of robotics, real-time dense geometric acquisition enables informative environment perception and serves as a valuable cue for localization and navigation. Besides, dense 3D models portrait scenes and produce insightful visualizations.

When we refer to 3D reconstruction, it is inevitable to consider the geometric representation. In the context of real-time reconstruction using consumer level sensors, the data structures that are robust to noise and suitable for data fusion are preferred. Therefore volumetric scalar fields (\eg signed distance field) have gained their reputation for the ability to easily integrate noisy data at various viewpoints; point-based methods are also appreciated for their elegance in math using filtering techniques. Mesh, as a widely-used classical 3D representation, however, is not paid much attention to for its loose organization of vertex arrays and their indices interpreted as triangles.

\begin{figure}[t]
	\begin{center}
		\includegraphics[width=1\linewidth]{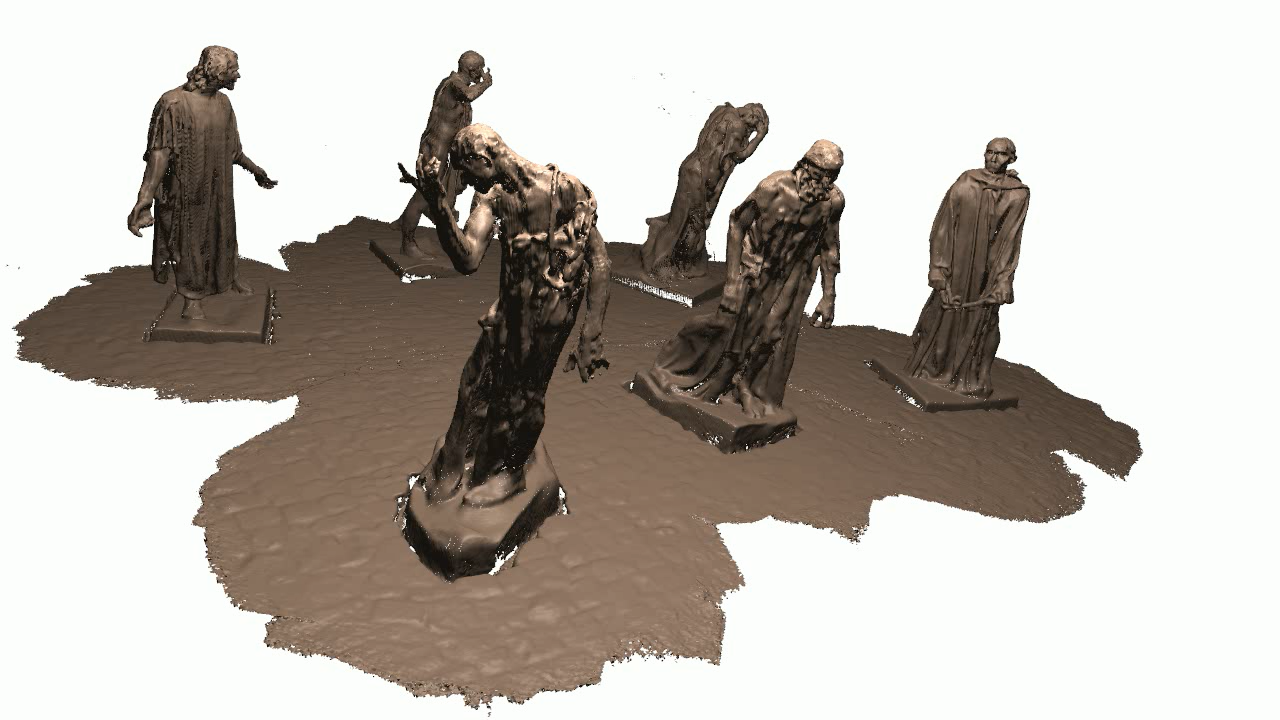} 
		\includegraphics[width=1\linewidth]{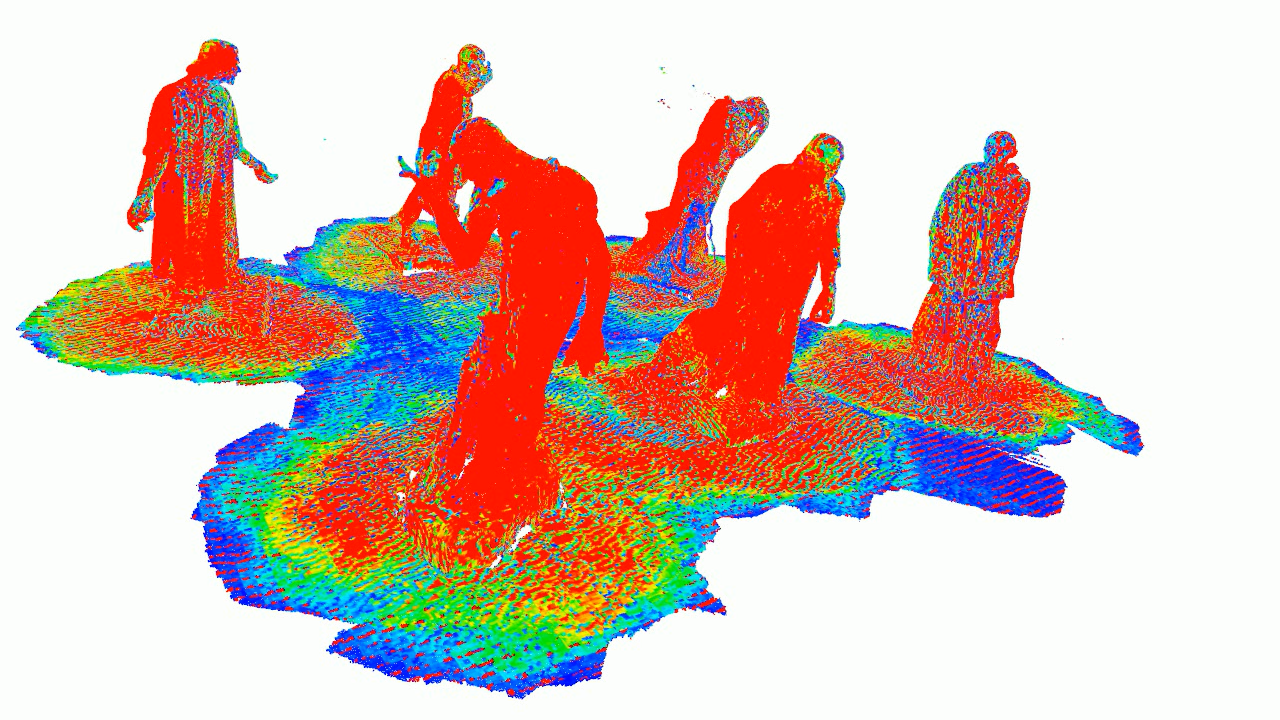} 
	\end{center}
	\caption{{\it Top}, final reconstructed mesh of scene {\it burghers} (cube resolution 8mm) rendered with Phong shading. {\it Bottom}, visualization of the duration of each vertex maintained in memory; the warmer the color, the longer the duration, indicating a stronger temporal consistency.}
	\label{fig:intro}
\end{figure}

Although in many cases not as suitable as other methods for real-time data fusion, mesh owns various advantages. Composed by triangles, it is highly efficient for rendering, acting as the default structure on most graphics hardwares and industrial softwares. Besides, it is a reasonable simplification and sampling of the continuous 3D surfaces that can provide control points especially useful for deformation estimation, essential in real-time dynamic reconstruction \cite{Newcombe2015}. Topology is reserved viewing the connectivity of vertices, hence 3D segmentation over mesh is also desirable to provide high-level understanding of the scene during data fusion \cite{Salas-Moreno2013}.

In view of this, mesh is extracted in many incremental reconstruction frameworks where it is indispensable. These implementations are, however, usually either functionally separated as utilities \cite{whelan2012kintinuous, InfiniTAMv3, Salas-Moreno2013, Newcombe2015}, or adopting loose mesh storage strategies, not fully taking the advantage of compact spatial representations \cite{Steinbrucker2014,Klingensmith2015}. This would impair the neatness of a reconstruction pipeline, possibly cutting off relations between mesh and latent data; duplicate vertices are prone to be allocated, losing the mutual connections between triangles. Besides, an additional data structure such as KD-Tree or octree is required if spatial vertex querying is needed, which is not uncommon in neighbor searching and model resampling.


In this paper we design an incremental mesh generation framework based on volumetric data structure using spatial hashing \cite{Nießner2013,Klingensmith2015}. Our major contributions include:
\begin{itemize}
\item A compact data structure that embeds the vertices in the volumetric grids. We utilize the one-to-one correspondences between voxel edges and mesh vertices, eliminating vertex redundancies.
\item A parallel mesh generation pipeline with online mesh extraction,  update, and garbage collection, linking the mature volumetric data fusion techniques \cite{Newcombe2011} and the mesh extraction functions available on volumes \cite{Lorensen1987}.
\item A simple yet effective mesh topology refinement algorithm. We reveal the deficiency in the mesh created by prevalent real-time volumetric scene reconstruction systems, and improve the triangle shape consistency by local shape regularization.
\end{itemize}

\section{Related Work}
{\bf 3D data representation for online fusion.}
Real-time dense 3D reconstruction of ordinary scenes requires data fusion, which is aimed at integrating data acquired at different viewpoints with possible overlaps, and reducing noise. To meet such demand, many representations have been proposed. A fairly popular strategy is to divide the world volumetrically, and analyze per-voxel local geometric information. Curless and Levoy \cite{Curless1996} introduce the signed distance function (SDF) to describe the Euclidean distance from each voxel center to its closest surface. Newcombe \etal \cite{Newcombe2011} adopt a truncated version of signed distance function (TSDF) and implement a real-time application on GPU which incrementally fuses depth data captured by a Kinect. Since it manages the spatial volume with a 3D plain array, its working space is limited due to memory constraints. Zeng \etal \cite{Zeng2013} utilize an octree to replace the plain array, reducing the memory consumption to some extent. Similarly, Steinbr\"ucker \etal \cite{Steinbrucker2014} propose an octree-based structure that is able to run in real-time on a CPU. Whelan \etal \cite{whelan2012kintinuous,Whelan2015} instead maintain a moving volume of active area, and generate mesh when a region is streamed out. Nie{\ss}ner \etal \cite{Nießner2013} use a 2-level cascade voxel hashing strategy to manage voxels that is highly efficient for GPU. The method is extended to CPU by Klingensmith \etal \cite{Klingensmith2015}, and is further optimized by K{\"a}hler \etal \cite{Kahler15}. Other than volumetric approaches, there are also point-based \cite{Keller2013} and surfel-based \cite{Kolev2014} methods to perform data integration. Marton \etal \cite{Marton2009} demonstrate an adaptive mesh generation method by directly re-sampling over point clouds, but the underlying KD-tree is not efficient enough to support real-time processing. Zienkiewicz \etal \cite{Zienkiewicz2016} fuse data into mesh with non-local optimizations; presented as a 2.5D height map, occlusions can hardly be handled. 

{\bf Real-time rendering.}
Regarding the underneath representation of 3D data type, \ie, volume, point cloud, and mesh, several approaches have been raised to reveal the underlying 3D surfaces so as to render and visualize. Rendering mesh is trivial, as the modern graphics pipelines are mostly designed for triangles. For volumetric data, there are mainly two options available: generate mesh at the isosurfaces from the volumetric field with methods such as Marching Cubes (MC) \cite{Lorensen1987} and fall back to the regular triangle rendering \cite{Steinbrucker2014, Klingensmith2015}, or directly trace each ray from the pixels of a virtual camera to find the intrinsic physical properties, \ie, the surfaces laying on the zero-crossing set \cite{Newcombe2011, Nießner2013, Kahler15}, which is in theory the same as surface determination in MC. Point-based rendering has also been proposed for dense visualization of point clouds \cite{Botsch2003}, usually based on splatting. Due to the architecture of modern graphics hardwares, the techniques other than mesh rendering are relatively more expensive and less compatible, therefore a conversion into mesh is preferred in various systems.

{\bf Mesh generation from volumes.}
The cornerstone of mesh generation from volumetric data is laid by Lorensen and Cline \cite{Lorensen1987} with MC. This simple algorithm that can run in parallel has been widely used up to now with various refinements \cite{newman2006survey,Zirr2011}. However, it is generally suitable for static data. In the real-time reconstruction systems, MC is usually implemented in its original form with minor adaptations to the data structure of the volumetric scalar field. Steinbr\"ucker \etal \cite{Steinbrucker2014} manage the 3D space with an octree and store mesh in each node with $8\times8\times8$ voxels, where complicated border situations are decided and a recursive search through the tree is processed. Klingensmith \etal \cite{Klingensmith2015} follow \cite{Nießner2013} and divide the space into spatial-hashed bricks, each holding a batch of voxels (\eg, 8$\times$8$\times$8). Only bricks in the sensor's viewing frustum will be operated for mesh generation, where a vector of mesh triangles are loosely maintained per brick. These triangles are not connected even with shared vertices; the incremental meshing for each frame can be described as an entire new mesh generation in local areas, where no temporal continuity is reserved.

\begin{figure}[t]
	\begin{center}		
		\includegraphics[width=0.95\linewidth]{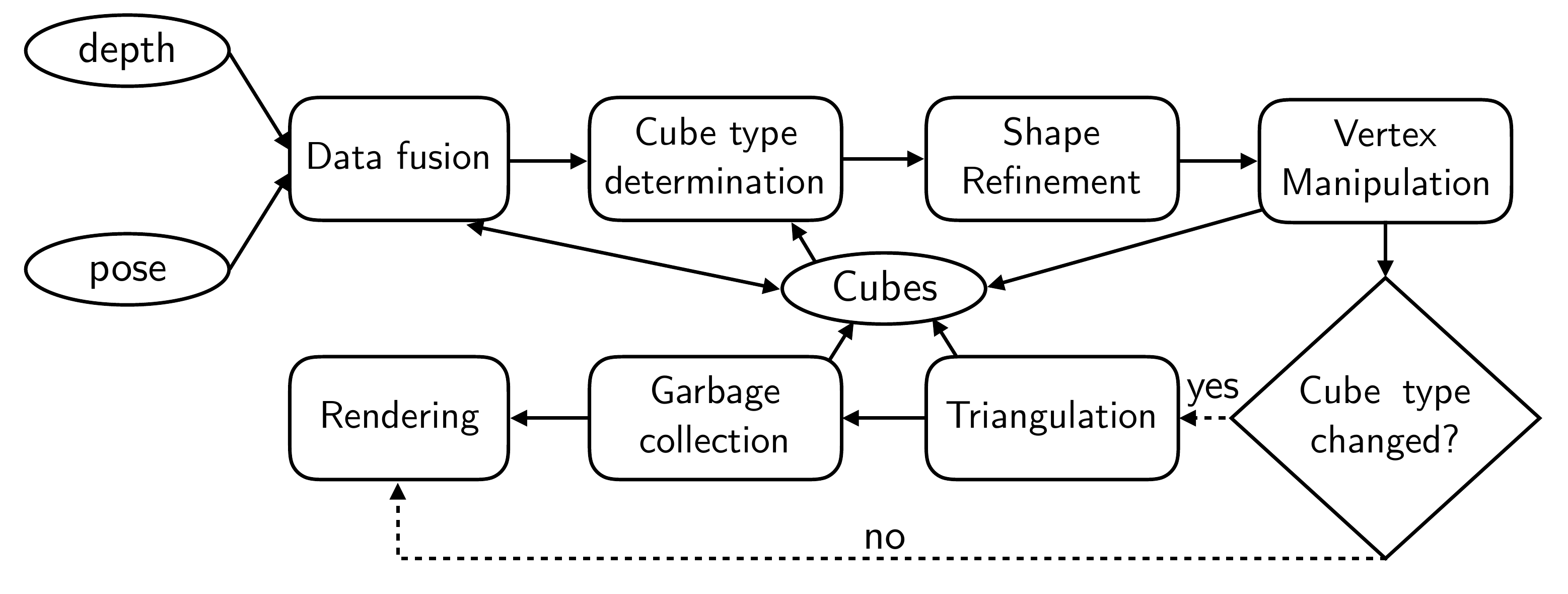}
	\end{center}
	\caption{System pipeline.}
	\label{fig:pipeline}
\end{figure}

\section{System Overview}
Our system extends the prevalent volumetric dense reconstruction pipeline in \cite{Nießner2013,Klingensmith2015}, illustrated in Fig.\ref{fig:pipeline}. The system is fed by a stream of depth images acquired by a hand-held sensor such as Kinect, along with the sensor's poses assumed known. In our work, the poses come from ground truth; RGB-D version of ORB-SLAM2 \cite{murORB2} can also be utilized as a black box for pose estimation. At each timestamp, the depth image is fused into the maintained volumetric {\it cube} field by changing TSDF distributions at the {\it corner} of each cube, to be discussed in \S\ref{sec:ds}; afterwards, a local mesh generation or update is performed based on MC \cite{Lorensen1987}, possibly accompanied by local refinements; finally mesh is reformatted for traditional triangle rendering instead of ray-casting. Our algorithms are specifically designed for parallel running on a GPU, but can be easily moved to CPU. In the following sections, the stages of the pipeline will be discussed one-by-one, except for mesh refinement and rendering. The former requires a detailed observation, therefore is separated, while the latter is too trivial to be discussed.

\begin{figure}[t]
	\begin{center}
		\includegraphics[width=0.8\linewidth]{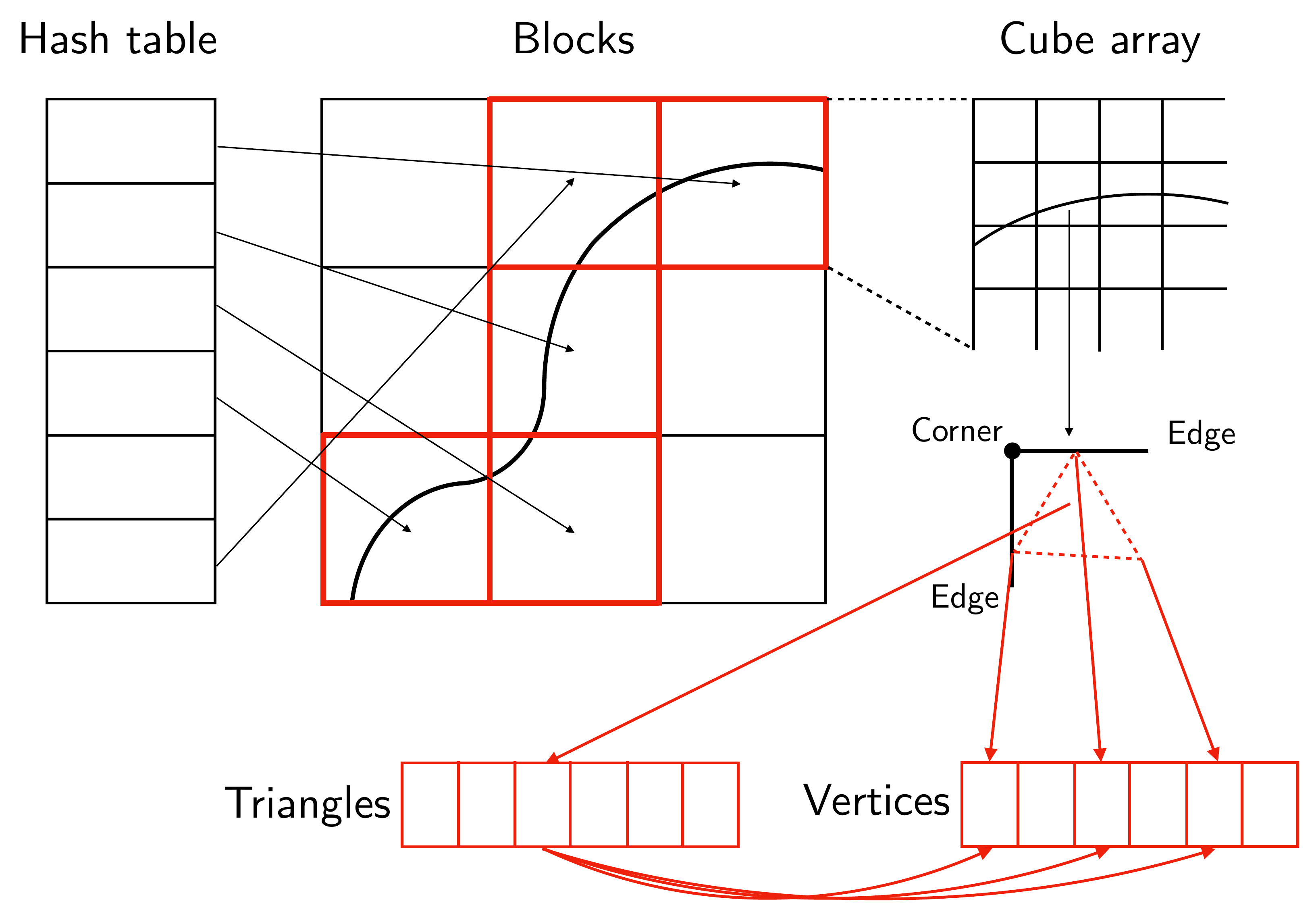}
	\end{center}
	\caption{2 level spatial management. Space is first coarsely divided into blocks around surfaces, managed by a hash table. Blocks are then further split into an array of cubes. Each cube holds 1 corner, 3 edges, and up to 5 triangles, which are allocated on the memory heap. Visualized in 2D for simplicity.}
	\label{fig:cascade}
\end{figure}

Fig.\ref{fig:cascade} shows a 2-level cascade structure to manage spatial information following \cite{Nießner2013}. At first, it splits the space into large {\it blocks} as the basic unit for spatial hashing; each block is further divided into many (\eg $8\times8\times8$) small {\it cubes} to hold local geometric information. This strategy constraints the size of hash table and therefore avoids hash collision to some extent, meanwhile guarantees the resolution of geometric information.

In \cite{Steinbrucker2014,Klingensmith2015,Kahler15} scalar geometric values, \ie TSDF, are stored at the fine-scale {\it voxel} level, while triangles are coarsely managed in the {\it block} level in hierarchy. Instead, we carefully maintain a {\it cube} structure to hold both triangles with their vertices and TSDF values, shown in Fig.\ref{fig:cube}; all the mesh manipulations are performed at the fine scale.

\section{Volumetric Triangle Representation} \label{sec:ds}
Terms are first introduced in this section. The space is split into {\it blocks} allocated only around object surfaces. A block is further divided into {\it cubes}, typically owning 8 {\it corners} and 12 {\it edges}; considering the overlap, however, only 1 corner and 3 edges need to be stored in average.

\begin{figure}[t]
	\begin{center}
		\includegraphics[width=0.8\linewidth]{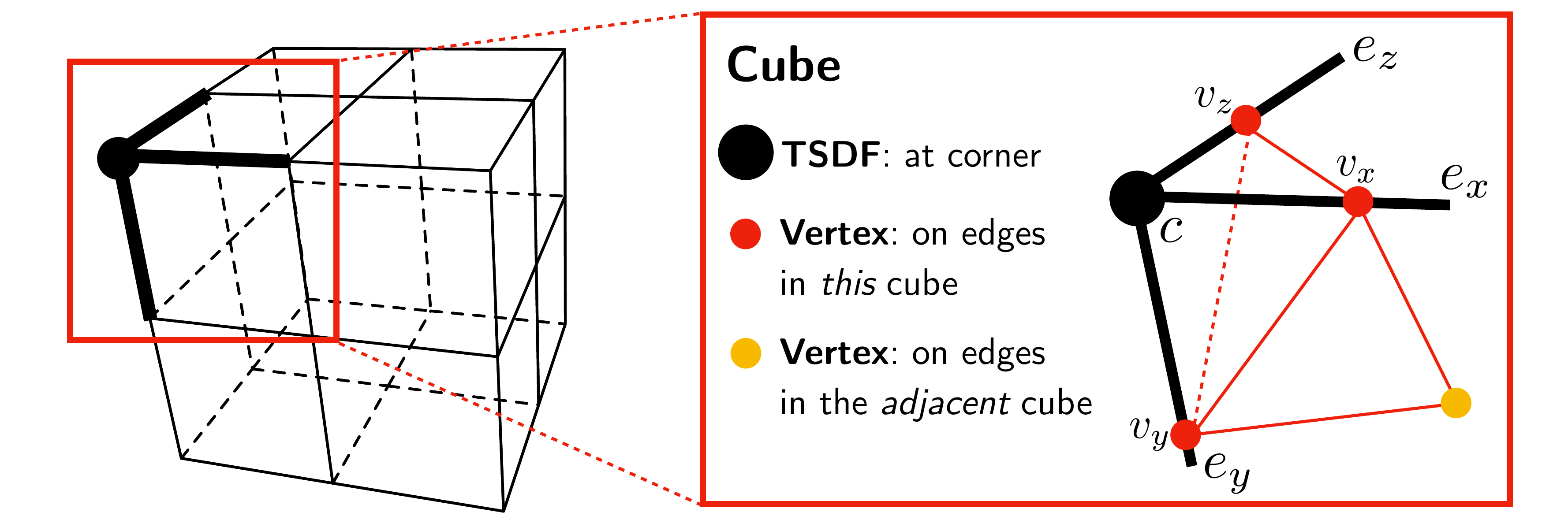}
	\end{center}
	\caption{Cube level data structure. Only 1 corner and 3 edges are maintained per cube, while others can be accessed at adjacent cubes.}
	\label{fig:cube}
\end{figure}

As illustrated in Fig.\ref{fig:cube}, we align each cube to the $xyz$ axises, maintain the corner at $c = (x_0, y_0, z_0)$, and preserve the edges $e_x = (l,0,0)$, $e_y = (0,l,0)$, and $e_z = (0,0,l)$ that start from $c$, where $l$ is the cube's side length. TSDF $d(c)$ is incrementally updated on $c$ for data fusion, and vertices $v_x, v_y, v_z$ intersected on axises, if existing, are stored on the correspondent edges with limited local degree of freedom ensured by MC. This binds the continuous vertex position to a discrete edge coordinate, making it possible for vertices to be directly accessed in $O(1)$ with a hash table visiting plus a local indexing; vertex sharing between adjacent cubes becomes especially simple via edge indexing. A cube also holds up to 5 triangles that connect the vertices on edges, which might come from a nearby cube; cube type in MC is recorded as a supplement to indicate the number and shape of triangles, both previously and currently, to be discussed in \S\ref{sec:cubetype}.

In terms of memory efficiency, edges and triangles in a cube are stored in pointer arrays of the size 3 and 5, while the pointed data are managed on the memory heap. This can be further optimized by saving only 1 pointer each for edges and triangles, where pointer arrays are also dynamically managed on the memory heap. 
%
A vertex stores position and normal, and reserves the space for color. In addition, we introduce a reference count to determine whether recycle is needed, referring to \S\ref{sec:garbage}. A triangle holds 3 pointers to index its vertices.

\section{Mesh Generation, Update, and Refinement}\label{sec:meshing}
\subsection{Block Collection and Data Fusion}\label{sec:fusion}
When a depth image $\mathcal{D}_i: \mathbb{R}^2 \to \mathbb{R}$~ along with a sensor pose (from sensor to world) ${}_s^wT_i = [{}_s^wR_i \mid {}_s^wt_i] \in SE(3)$ is received at the timestamp $i$, we first find each valid pixel $p \in \Omega_i \subset \mathbb{R}^2$, where $\Omega_i$ is the set of valid pixels in $\mathcal{D}_i$, and form a ray:
\begin{equation}
r = {}_s^wt_i + \lambda {}_s^wR_i \mathcal{D}_i(p) K^{-1}\tilde{p},
\end{equation}
where $\tilde{p}$ is the 3D homogeneous coordinate, $K$ is the intrinsic matrix of the sensor, $\lambda$ is the length parameter along the ray, and $\mathcal{D}_i(p)$ reads the depth value at $p$. In a certain range around the scanned point along $r$, \ie $\lambda \in [1-\delta, 1+\delta]$ , blocks are collected and will be allocated if not already done. Therefore only the blocks affected by new observations will be processed.

After collection, every corner of cubes $c \in \mathbb{R}^3$ inside the gathered blocks are projected to the depth image to find the approximately closest scanned point, and truncated distance is computed accordingly:
\begin{align}
\tilde{d}_i(c) &= \phi(\mathcal{D}_i(K{}_s^wT_i^{-1} c) - ({}_s^wT_i^{-1}c).z),
\end{align}
where $\phi(\cdot)$ is the truncation function and $({}_s^wT_i^{-1}c).z$ is the depth of $c$ in the camera coordinate system; $\tilde{d}_i(c)$ is then integrated into stored $d(c)$, details discussed in \cite{Newcombe2011}. TSDF value inside the sensor's viewing frustum around surfaces will be updated, being the basis of mesh generation. 

\subsection{Cube Type Determination} \label{sec:cubetype}
MC \cite{Lorensen1987} is utilized in the generation of mesh in the following sections. In MC, a table $\mathcal{T}: \{0,1\}^8 \to \{0,1\}^{12}$ is precomputed to indicate the triangle distributions, \ie the number of triangles and on which edges do their vertices lie. Each bit of $t$ denotes whether the scalar value at the related corner (in our case, $d(c)$) is below an isovalue (in our case, 0); each bit of $\mathcal{T}(t)$ indicates the existence of a vertex on the corresponding edge, 3 in the current cube and 9 in adjacent cubes. In most situations, the access of scalar value at corners is as trivial as visiting an adjacent value in a plain array. There exists border cases that the neighbor cube providing shared corner is not in the same block, where an additional $O(1)$ spatial hash table lookup is required. The current cube type $t_i$ is computed and stored along with the previous cube type $t_{i-1}$ to provide a cue for temporal consistency.

\subsection{Vertex Initialization and Update} \label{sec:vertex-sharing}
Having determined $\mathcal{T}(t)$, linear interpolation of endpoints' positions of an edge whose indicator bit is 1 will be computed in order to decide the position of the vertex it binds. The assignment is lazy: vertices are initialized only when first used; otherwise an update is sufficient.

The most elaborate part of this method different from the original version lies in vertex sharing in the neighbor cubes. In a serial implementation, \eg loop based CPU version, this is trivial once we choose the correct loop order. This is however, absolutely non-trivial when the program runs on a GPU where thousands of stream-processors are working simultaneously and vertices are determined in parallel. If no care is taken of, memory leak will be severe, causing 2 to 3 times of additional memory consumption; unexpected results may also take place. We attempt two solutions to guarantee the correctness of sharing:

\begin{figure}[t]
	\begin{center}
		\includegraphics[width=0.5\linewidth]{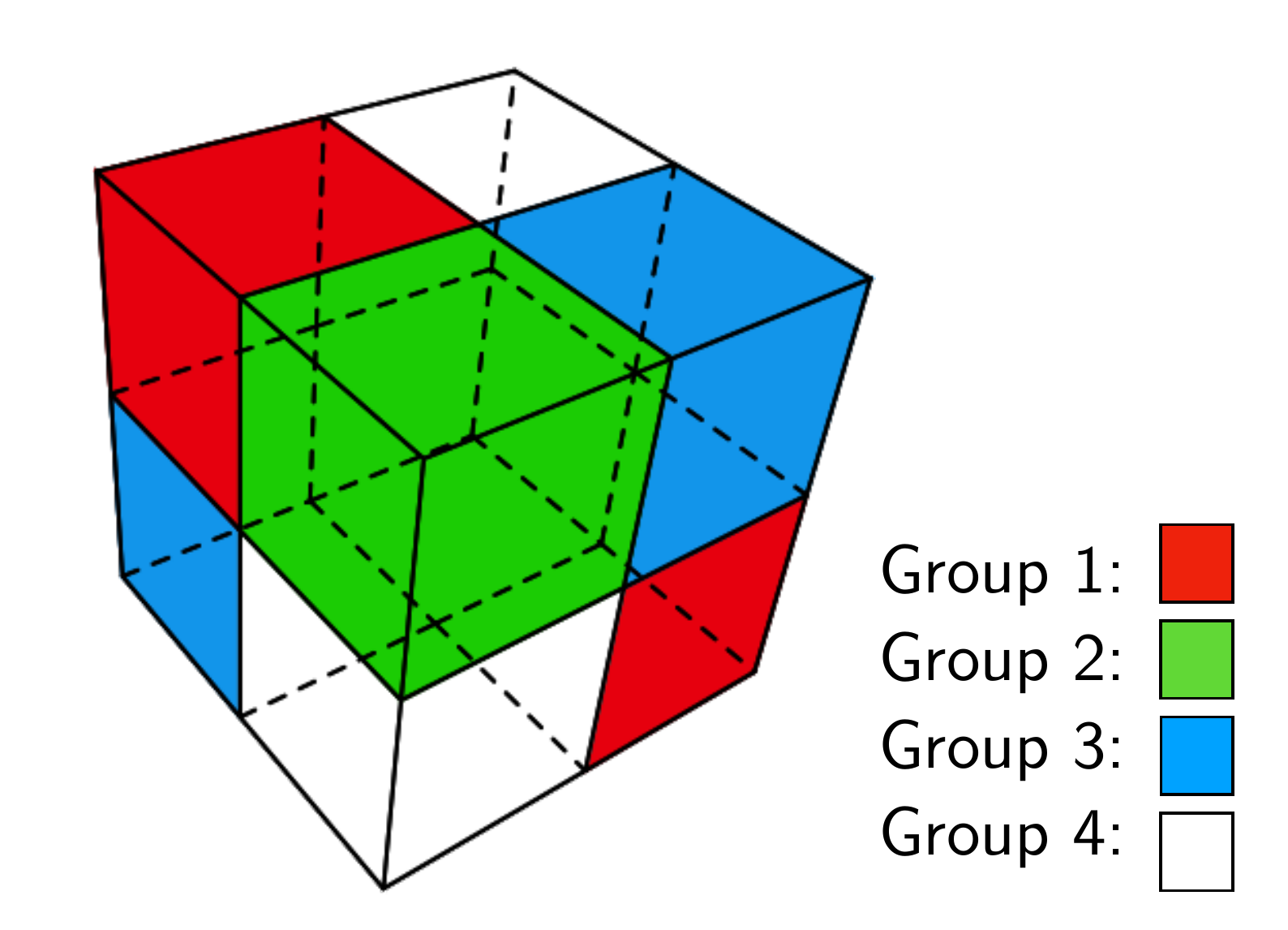}
	\end{center}
	\caption{The grouping of cubes. Cubes at opposite poles are gathered with no overlapping edge hence no corresponding vertex. }
	\label{fig:reduction}
\end{figure}

{\bf Lock-free version.}
A typical method to avoid conflict between threads is to utilize the reduction method with a divide-and-conquer strategy \cite{harris2007optimizing}. In its original form to sum up an array of numbers, the array is divided into two non-overlapping parts and summed up in each part; the process is iteratively operated until the array is not separable.

Inspired by this manipulation, we divide the 3D array of cubes inside a block into several groups in which no overlap exists. Illustrated in Fig.\ref{fig:reduction}, we divide a $2\times2\times2$ cube into 4 parts, which can also be extended to a wider region. As no edges are shared, this process is lock-free and can run in parallel.

{\bf Lock-based version.}
Lock is another traditional solution for resource sharing. A pure mutex-based operation will be inappropriate, however, as thousands of threads querying mutexes will easily lead to severe deterioration of performance. Instead, we adopt an atomic operation under such circumstances. In this implementation, only the first thread who atomically acquires a vertex will have the privilege to allocate and assign it. Since the interpolation ratio of an edge's endpoints are already determined in the data fusion stage, the correctness of the vertex's position will hold for the other threads.

\subsection{Triangulation}
Up to this stage, we have determined the vertices of triangles to be processed. To reduce the cost of triangle allocation and assignment, we compare $t_i$ and $t_{i-1}$: if they coincide, common in the incremental process, the list of triangles and their vertices will remain unchanged inside the cube, keeping a temporal consistency; otherwise the previous triangles will be cleared and new ones will be created.

\begin{figure}[t]
	\begin{center}		
		\includegraphics[width=1.0\linewidth]{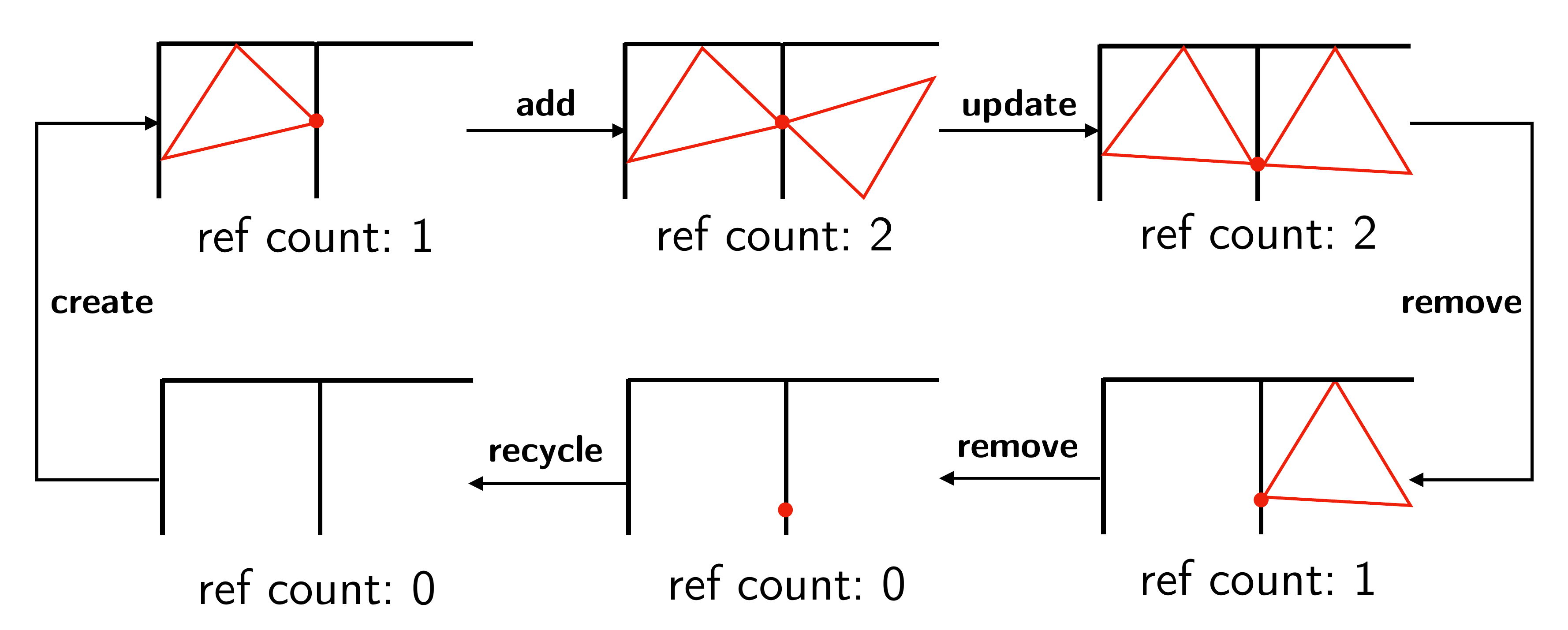}
	\end{center}
	\caption{Change of a vertex's reference count according to triangle insertion and deletion. Recycling will be triggered when a vertex is not referenced anymore.}
	\label{fig:refcount}
\end{figure}

\subsection{Garbage Collection} \label{sec:garbage}
The shared vertices are referenced by and only by triangles. In order to manage memory correctly, we use the reference counting technique. When a new triangle is created, the reference count of its related vertices will be increased by 1; when a destroying operation takes place, a symmetry decrease operation will be processed, as Fig.\ref{fig:refcount} illustrates. The vertices with a 0 reference count will be regarded as garbage and recycled, waiting for a new allocation. 

Aside from a recycling indicator, the reference count can also be regarded as the degree of a vertex in topology, which might serve as a useful property in mesh analysis. 

\begin{figure}[t]
	\subfloat[] {
		\includegraphics[width=0.55\linewidth]{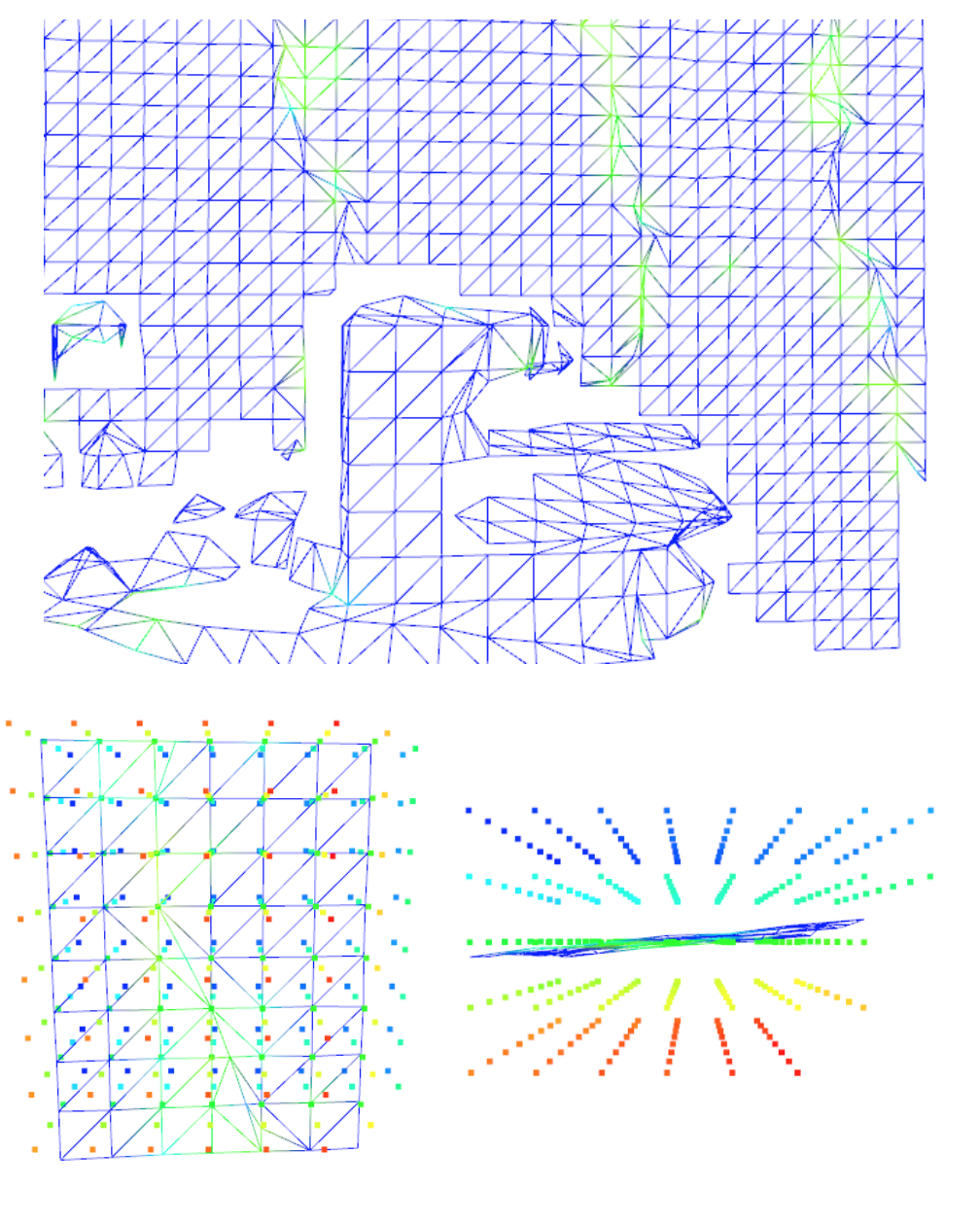}
		\label{fig:ambiguous-a}
	}
	\subfloat[] {				\includegraphics[width=0.32\linewidth]{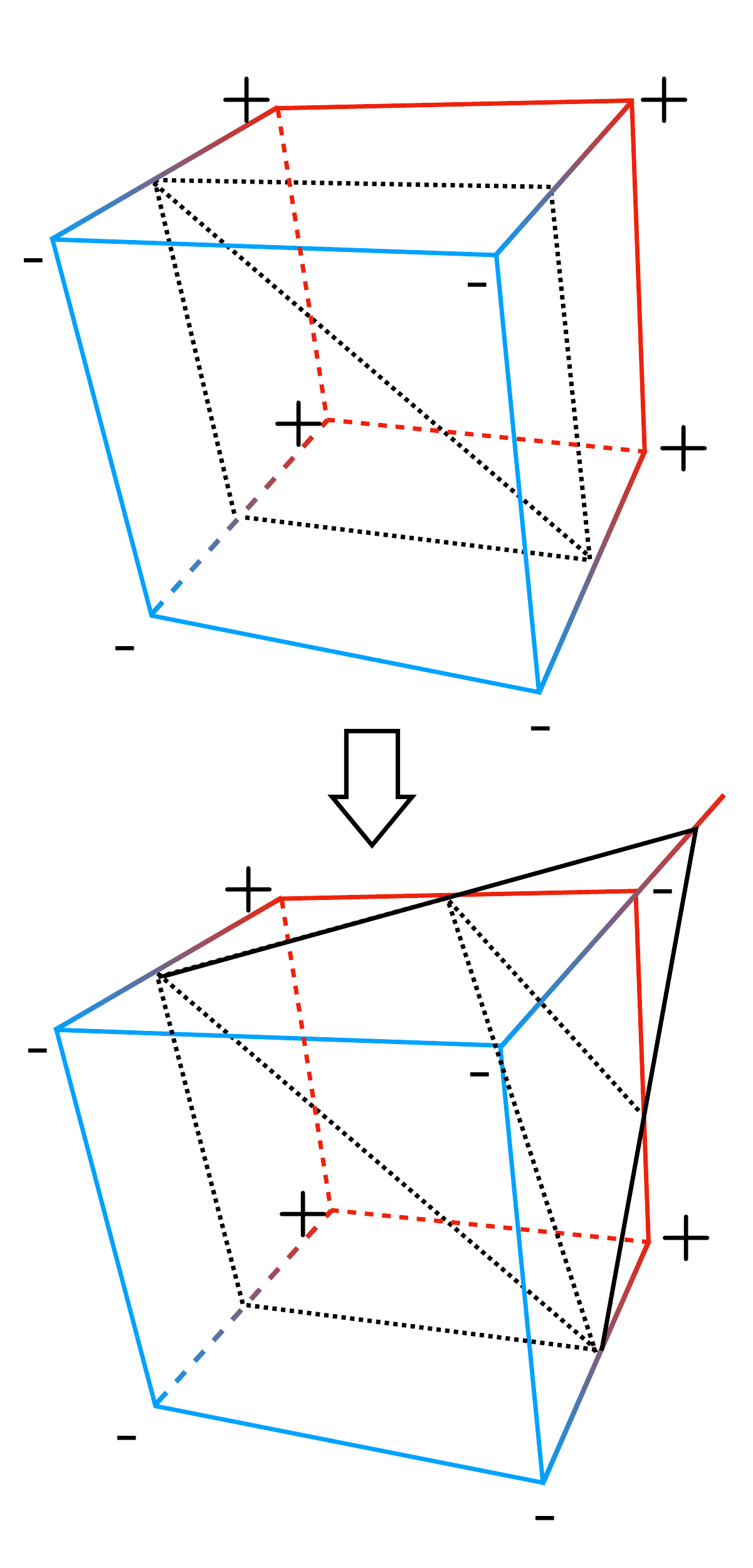}	
		\label{fig:ambiguous-b}
	}

	\caption{Illustration of `cracks'. {\it (a), Top}, a general look into the {\it copyroom} scene where a printer is placed in front of a wall. Color of mesh represents the confidence of TSDF values around vertices, green accounts for more unsteadiness. Note the irregular green triangles appear in pattern. {\it Left bottom}, a closer view of a specific block, where TSDF values at corners are also visualized in color. The warmer the color, the more positive the value; green is approximately 0. {\it Right bottom}, upper view of the block with mesh. Triangles whose vertices are around zero-value TSDF corners are prone to be irregular. {\it (b)}, abruptly changed triangles in {\it dotted lines} emerge due to disturbance of TSDF value at one corner. }
	\label{fig:ambiguous}
\end{figure}

\subsection{Mesh Refinement}\label{mesh-refine}
\begin{figure}[t]
	\begin{center}
		\includegraphics[width=0.9\linewidth]{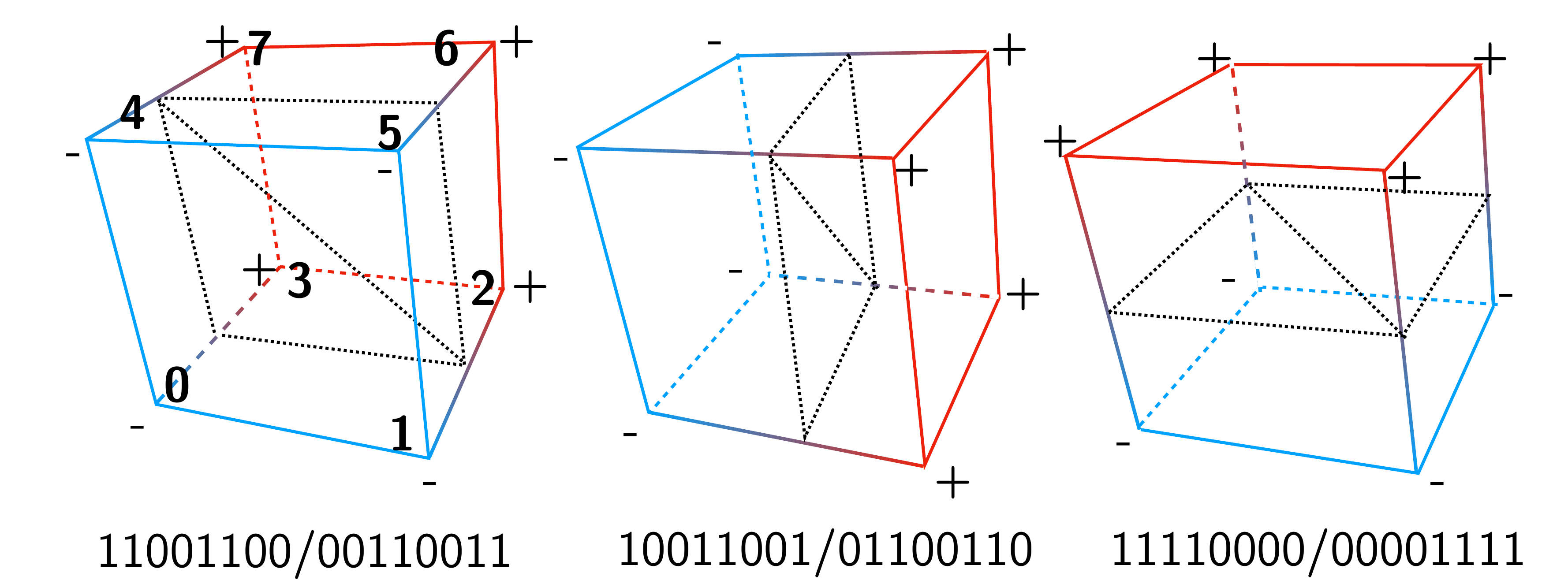}
	\end{center}
	\caption{Regular and most frequently created triangle shapes in ordinary scenes. The order of 8-bits is presented in the first cube.}
	\label{fig:hamming}
\end{figure}

A problem of mesh extracted from volumetric fields during online data fusion can be found in the illustrated figures in literature \cite{Steinbrucker2014,Klingensmith2015}, and also appears in our system. It is an interesting phenomenon depicted in Fig.\ref{fig:ambiguous-a}, where irregular triangles appear in an observable pattern. After a careful analysis, we find it is the limited resolution and the principle of MC that leads to the deficiency.
In most cases, a real world plane will go through the middle of a cube, see Fig.\ref{fig:hamming}; the corners of the cube at two sides separated by the plane will hold TSDFs who are dominated by a series of $\tilde{d}_i(c)$ (see \S\ref{sec:fusion}) of the same sign. In such a case, the cube type $t$ can be determined with confidence, producing two triangles that is neat enough to represent the crossing plane of a cube. However, when a plane in the world coordinate system is not strictly parallel to the axises of cubes, it is highly possible to intersect cubes at corners, as illustrated in Fig.\ref{fig:ambiguous}. In such cases, positive and negative $\tilde{d}_i(c)$ are distributed evenly at $c$; a small disturbance would lead to the flip of sign of $d(c)$, hence the bit-array $t$ will be directly affected, resulting in an abrupt change of the output $\mathcal{T}(t)$ and its correspondent triangle distribution shown in Fig.\ref{fig:ambiguous-b}. This event would repeat itself along the plane every time such an intersection occurs.

Attene \etal \cite{Attene13} provided a comprehensive review of available mesh repair techniques in the application perspective, yet a satisfying on-the-fly solution does not appear regarding online mesh generation. Dzitsuik \etal \cite{Dzitsiuk2017} have came up with the idea of fitting planes to increase the smoothness in an incremental fashion. This method runs efficiently when the scene is smooth with many planes, but might face performance problems in a complex environment with a high resolution according to our experiments.  In the context, we introduce an simple yet effective local method to reformat the triangle shapes. 

Intuitively, in the disturbed cube in Fig.\ref{fig:ambiguous-b}, if we extend the affected triangle edges, they would approximately intersect outside the cube (may not exactly intersect, but fairly close given a small TSDF's absolute value), forming a large triangle bounded by the {\it solid lines}. Assuming a smooth transition of TSDF in the local area, the extrapolated vertex would coincide with the vertex held by its correspondent neighbor. Therefore geometrically it is reasonable to eliminate the small fragments and maintain one large triangle instead. A tricky implementation is adopted to achieve the target: we simply set the type of the disturbed cube to the undisturbed type and run MC. The interpolation in MC would, with the same equation, serve as extrapolation given the same sign of TSDF of two adjacent corners.

The following operation detects the disturbance: given the 8-bit vector $t_{i-1}$ and $t_i$ denoting the previous and currently estimated cube type, if
\begin{align}
d_H(t_i, t_{i-1}) &\le 3,\label{equ:flip0}\\
d_H(t_i, t_{rj}) &\le 3, \exists j \in \{1 \cdots 6\},\\
|d(c_k)| &< \epsilon, \forall k \mid t_{i,k} \oplus t_{rj, k} = 1, \label{equ:flip2}
\end{align}
are satisfied, where $d_H$ denotes the Hamming distance, $t_{i,k}$ is the $k$th bit of $t_i$, $d(c_k)$ reads the TSDF value at the $k$th corner, $\oplus$ is the xor sign, $\epsilon$ is a preset threshold proportional to cube size, and $t_{r1...6} = \{11001100, 00110011, 10011001, 01100110, 11110000, \\00001111 \}$ hold the `regular cube' types shown in Fig.\ref{fig:hamming}, then we assume it is the disturbances at the $k$th corners in Eq.\ref{equ:flip2} that flip the sign, leading to irregularity. Under such circumstances, $t_i = t_{rj}$ is applied before MC.

This approach is robust: Hamming distance between each pair of regular type vector $t_{rj} \in t_{r\{1\cdots6\}}$ is either 4 (perpendicular) or 8 (parallel with all sign reversed), therefore choosing 3 as a discriminating value in Eq.\ref{equ:flip0}-\ref{equ:flip2}, the triangles will adhere to the closest regular type in Fig.\ref{fig:hamming}; the sign reversion will also be strongly limited by the temporal constraint $t_{i-1}$ and the tolerant threshold $\epsilon$.

\begin{figure}[t]
	\begin{center}
		\subfloat[] {
			\includegraphics[width=0.48\linewidth]{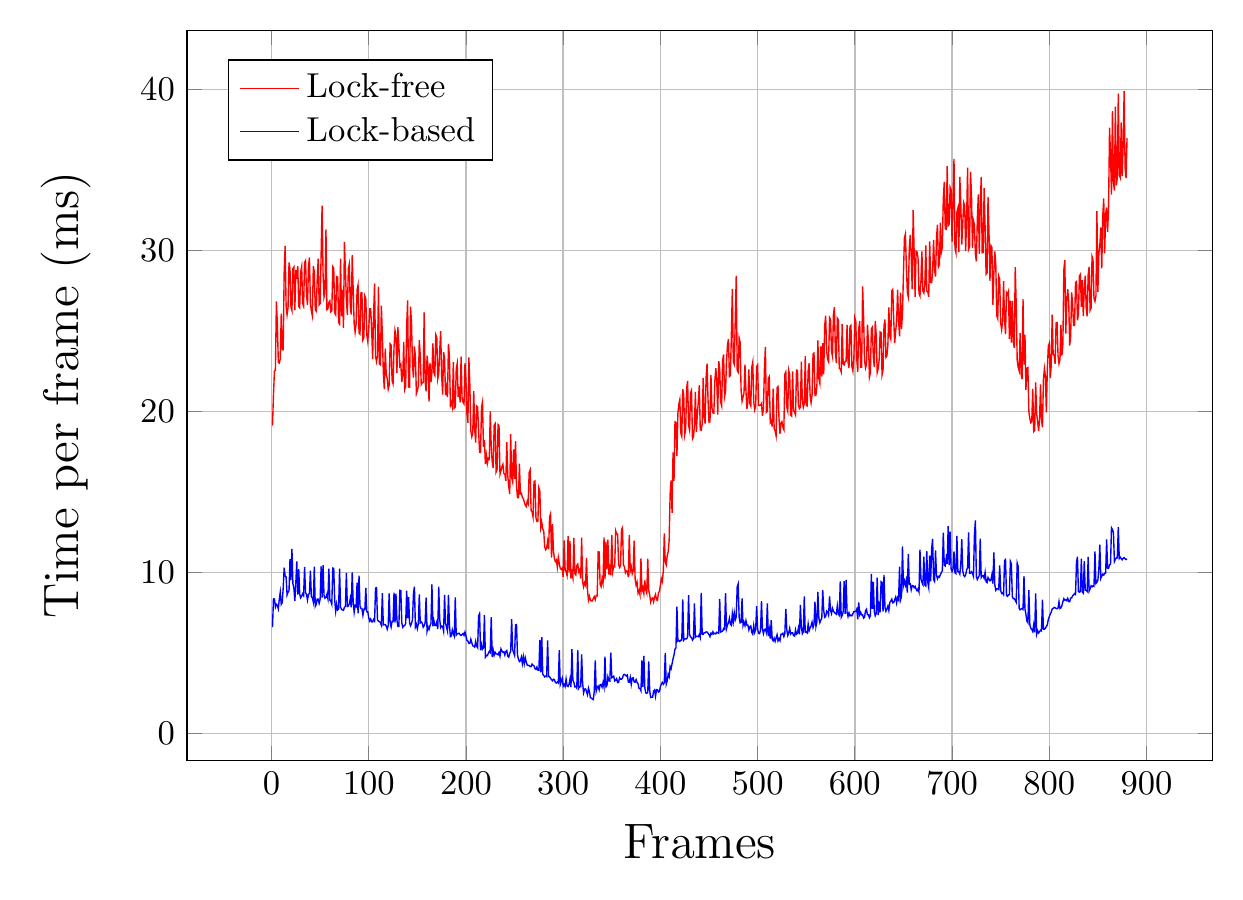}
			\label{fig:framewise-chart-a}
		}	
		\subfloat[] {
			\includegraphics[width=0.48\linewidth]{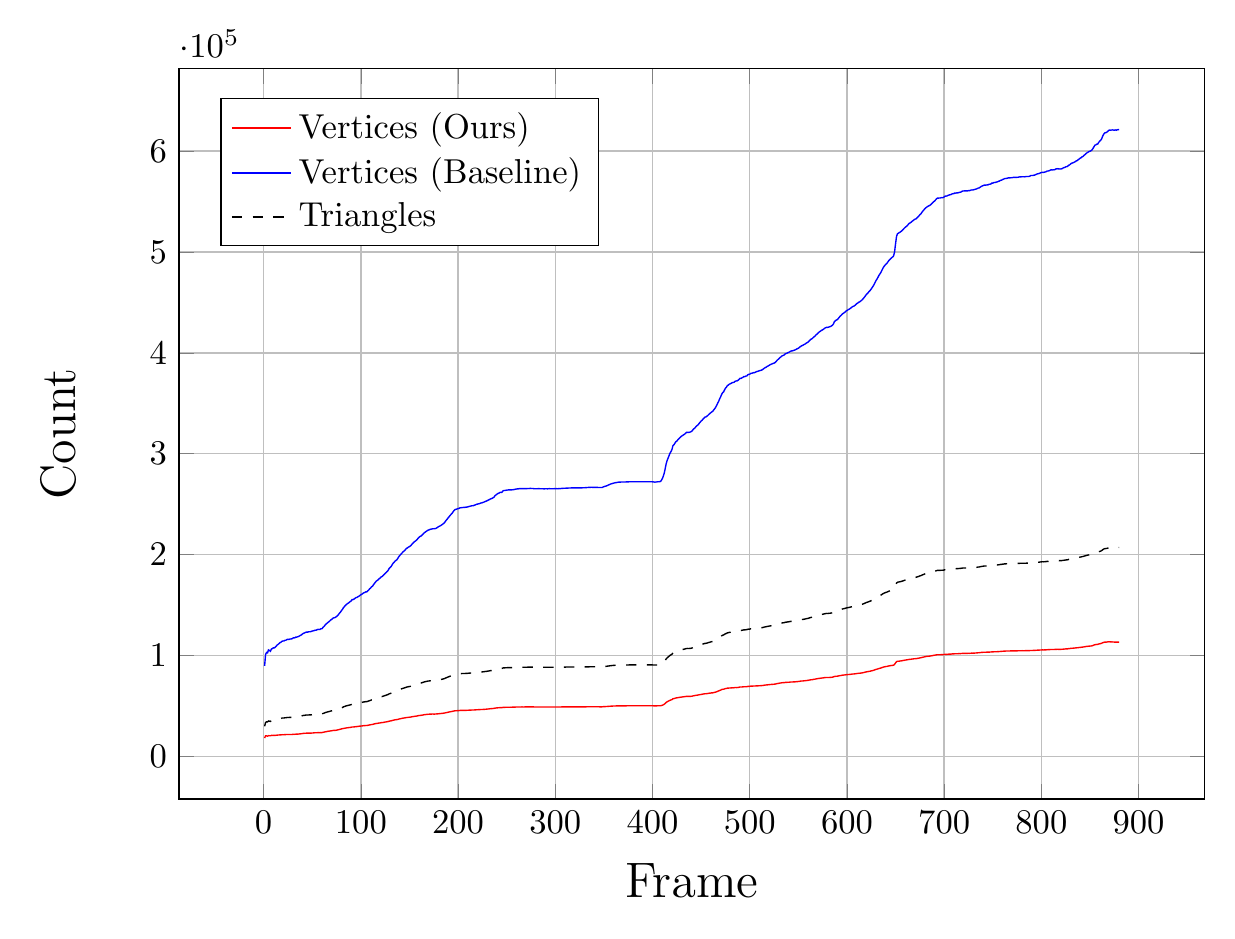}	
			\label{fig:framewise-chart-b}
		}
	\end{center}
	\caption{Experiment results of an incremental reconstruction on {\it office2} sequence with a 3cm cube resolution. {\it (a)}, running time comparison between {\it lock-based} and {\it lock-free}. {\it (b)}, mesh memory consumption.}
	\label{fig:framewise-chart}
\end{figure}

\section{Experiments}\label{sec:experiments}
We test our method on various RGB-D datasets, including ICL-NUIM \cite{handa2014}, TUM \cite{sturm12iros}, and datasets provided by Zhou and Koltun \cite{Zhou2013}, where depth images with registered poses are all provided. The experiments are conducted on a laptop with an Intel Core i7-6700HQ CPU, and an Nvidia GTX 1070M graphics card. We take advantage of the core components including GPU hash table and data fusion from the open-source code provided by \cite{Nießner2013}, and implement the meshing pipeline entirely. The code is written in C++, with CUDA 8.0 for parallel computation and OpenGL 3.3 for rendering. In CUDA, each {\it block} is assigned to a stream processor, and each {\it cube} is manipulated by a thread. In all configurations, a {\it block} contains $8\times8\times8$ {\it cubes}. The generated mesh is directly compressed and copied in GPU memory from the CUDA context to OpenGL for real-time feedback. Cube resolutions vary from 8mm (typically used for fine-grained scene reconstruction) to 3cm (usually set for global mapping in SLAM tasks).

Our pipeline is compared against \cite{Klingensmith2015} as the baseline, for which we also implemented a GPU version. For simplicity, all the mesh is stored in a global array instead of arrays allocated per block in \cite{Klingensmith2015}. Without loss of fairness, we generate mesh only for blocks in viewing frustum to test running speed, and for all blocks to test memory usage.

\subsection{{\it Lock-based} and {\it Lock-free} Comparison}
In \S\ref{sec:vertex-sharing} we have discussed two possible solutions for parallel vertex sharing. To determine which approach to adopt, we evaluate both and draw the conclusion that the {\it lock-free} implementation, although theoretically achievable, is not preferable to the \emph{lock-based} version.

Fig.\ref{fig:framewise-chart-a} illustrates the result of running time for  meshing stage of two methods. The time of {\it lock-free} is 2 to 3 times of the {\it lock-based} version; it seems that the avoidance of thread conflicts cannot compensate for the expenses of group-level serial launches. This trend also holds for higher cube resolutions. In view of this, we choose the {\it lock-based} version in following experiments. The idea of grouping cubes in {\it lock-free} might be utilized in a multi-threaded CPU version where the order of loop is critical.

\subsection{Memory and Running Time Results}
\begin{figure}[t]
	\begin{center}
		\subfloat[]{
			\includegraphics[width=0.48\linewidth]{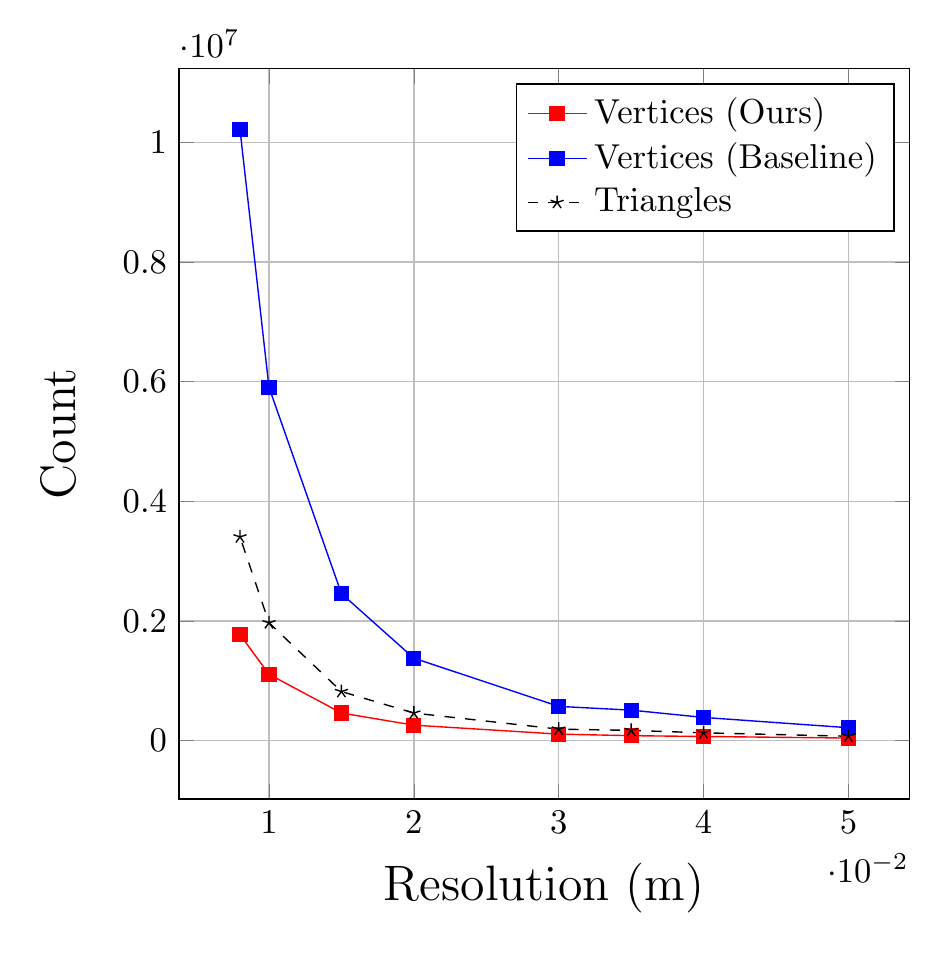}
			\label{fig:resowise-chart-a}
		}
		\subfloat[]{
			\includegraphics[width=0.48\linewidth]{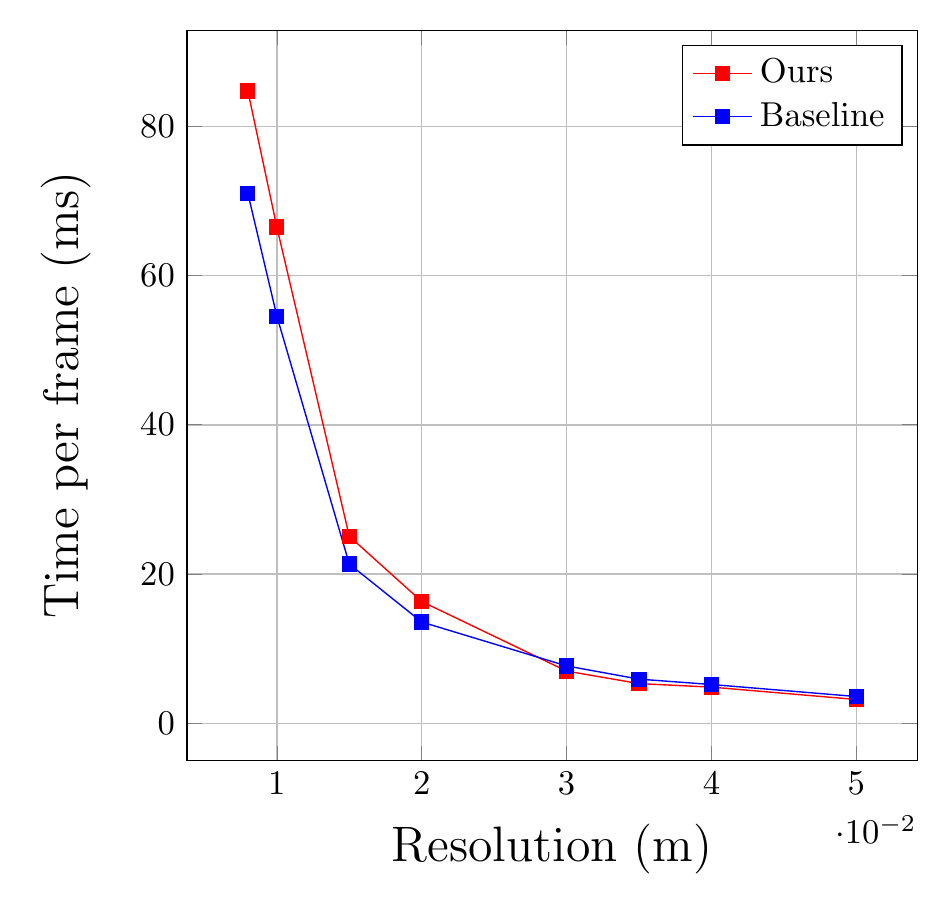}		\label{fig:resowise-chart-b}
		}
	\end{center}
	\caption{Comparison of our method and baseline. {\it (a)}, mesh memory consumption. {\it (b)}, average running time. Both experiments are conducted on the {\it office2} dataset with a series of resolution configurations.}
	\label{fig:resowise-chart}
\end{figure}

\textbf{Memory}. Experiments are first conducted to show reduction of mesh memory consumption. The number of triangles should be in theory identical for both the proposed method and the baseline; in experiments, there is a difference (generally $\le 5\%$) due to thread conflicts in hash table entry allocation \cite{Nießner2013}, and is ignored in figures for simplicity. We focus on vertex count, representing the geometry of scenes.

Fig.\ref{fig:framewise-chart-b} shows a typical trend of mesh accumulation during the sequential online reconstruction.  The growth of vertex count is significantly constrained in our method, compared to baseline. Fig.\ref{fig:resowise-chart-a} illustrates the number of vertices with different resolution selections. When the precision is fairly high, the gain will be considerable. In Table \ref{tab:table} we list the memory consumption for several datasets.

It is reasonable that vertices are even fewer than triangles. Consider a mesh that looks like the regular part in Fig.\ref{fig:ambiguous}: when we take an area of $w\times h$, the number of triangles will be $2wh$, while the number of vertices will be $(w+1)(h+1)$, hence in the infinite case the triangles will be $\lim_{w,h\to \infty} 2wh/[(w+1)(h+1)] = 2$ times of the vertices. Therefore, any vertex count that is greater than half of triangle count will be valid.

At current, although a reduction of vertex count is apparent, the memory cost in total (at 8mm cube resolution, 50000 blocks, 1.8M vertices, and 3.5M triangles, which is enough for all our test scenes) is in fact increased to about 1.6GB, since the data structure of a cube is not fully optimized, storing 56 bytes per unit. If it were minimized to 24 bytes using the techniques discussed in \S\ref{sec:ds}, the total memory including the cubes and mesh they hold will be reduced to around 700MB,  approximately the same as the memory of a TSDF field plus the non-optimized mesh \cite{Klingensmith2015}. With a similar total memory cost, our method reduces 3D model size, holds much more geometric information such as connectivity, and supports $O(1)$ vertex accessing.

\textbf{Time}. While introducing additional computations, we manage to maintain the running speed of meshing stages (refinement included) in general. For relatively low cube resolution, \eg 3cm, the running time is slightly faster than the baseline, as shown in Fig.\ref{fig:resowise-chart-b} and Table \ref{tab:table}. It turns out that our method becomes slower than baseline in the very dense case, \eg 8mm. This might be improved by dealing border cubes per block specifically, where many redundant hash queries are processed. In spite of this, our data structure serves as a trade-off between efficient mesh accessing and management, and fast mesh generation.

\begin{figure}[t]
	\begin{center}
		\subfloat[] {
			\begin{tabular}[b]{c} 
				\includegraphics[width=0.28\linewidth]{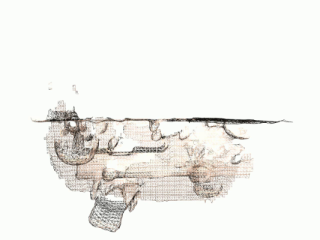}
				\includegraphics[width=0.28\linewidth]{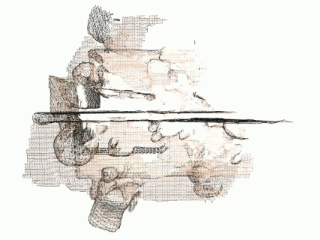}
				\includegraphics[width=0.28\linewidth]{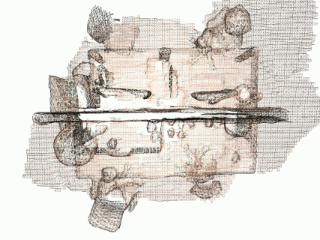}\\

				\includegraphics[width=0.28\linewidth]{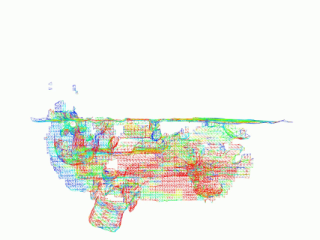}
				\includegraphics[width=0.28\linewidth]{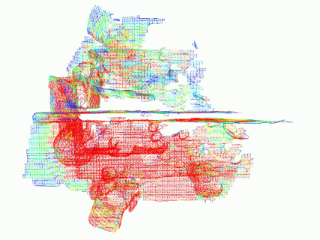}
				\includegraphics[width=0.28\linewidth]{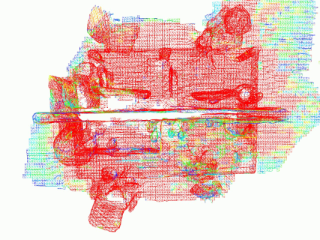}\\

				\includegraphics[width=0.28\linewidth]{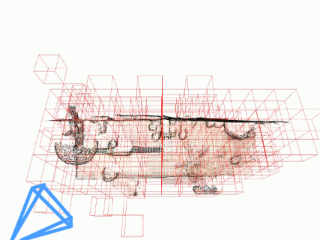}
				\includegraphics[width=0.28\linewidth]{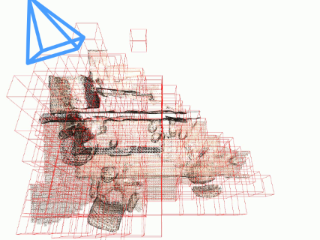}
				\includegraphics[width=0.28\linewidth]{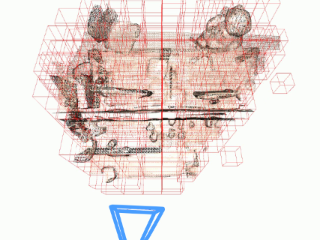}
			\end{tabular}
			\label{fig:incremental-a}
		}
		
		\subfloat[] {
			\begin{tabular}[b]{c} 
				\includegraphics[width=0.28\linewidth]{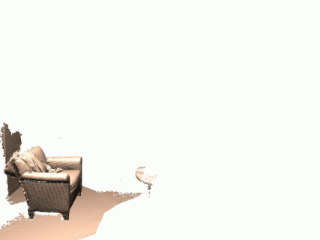}
				\includegraphics[width=0.28\linewidth]{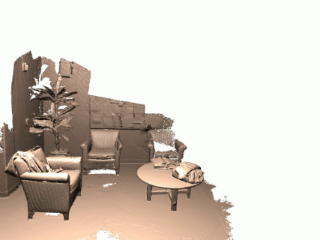}
				\includegraphics[width=0.28\linewidth]{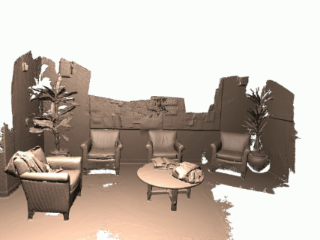}\\
				
				\includegraphics[width=0.28\linewidth]{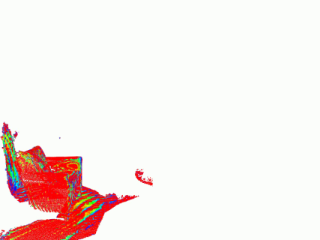}
				\includegraphics[width=0.28\linewidth]{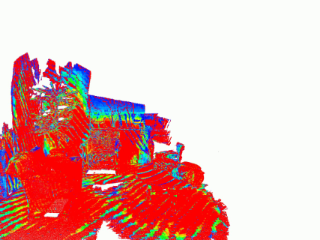}
				\includegraphics[width=0.28\linewidth]{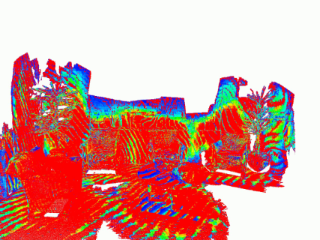}\\

				\includegraphics[width=0.28\linewidth]{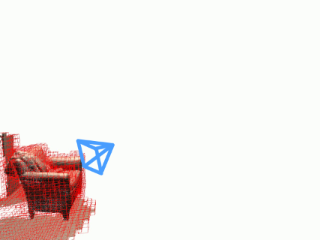}
				\includegraphics[width=0.28\linewidth]{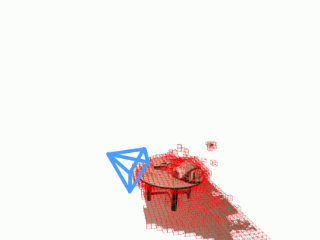}
				\includegraphics[width=0.28\linewidth]{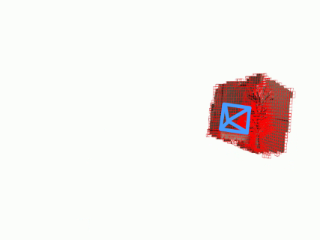}
			\end{tabular}
			\label{fig:incremental-b}
		}	
	\end{center}
	\caption{Incrementally reconstructed mesh. {\it(a)}, {\it household} with cube resolution 3cm and max scanning range 2.5m. {\it (b)}, {\it lounge} with cube resolution 8mm and max scanning range 1.6m (20 frames with heavy motion blur were manually filtered out). Each 3 rows from top to down: global mesh; visualized duration of vertices, where a warmer color indicates a longer sustained time; locally updated mesh in viewing frustums, where red bounding boxes represent blocks and blue pyramids denote frustums (enhanced for easier recognition). Best viewed in color and enlarged.}
	\label{fig:incremental}
\end{figure}

\begin{figure}[ht]
	\begin{center}
		\subfloat[] {
			\includegraphics[width=1.0\columnwidth]{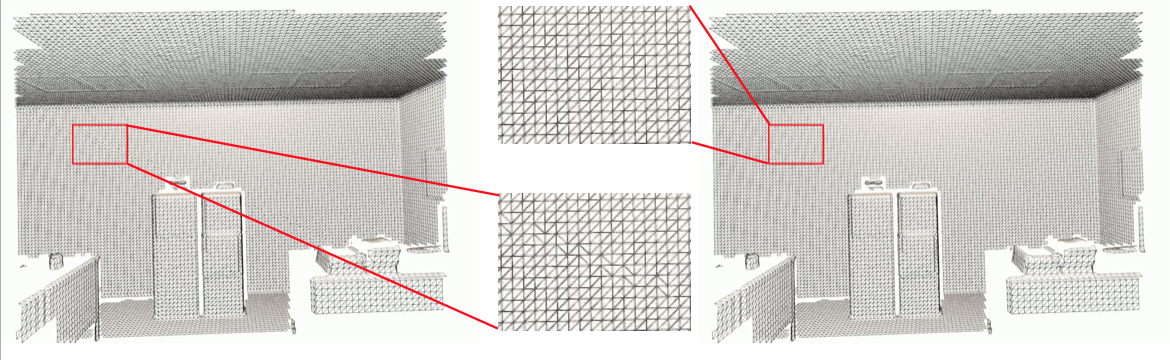}		
			\label{fig:refinement-a}
		}\\
		\subfloat[] {
			\includegraphics[width=1.0\linewidth]{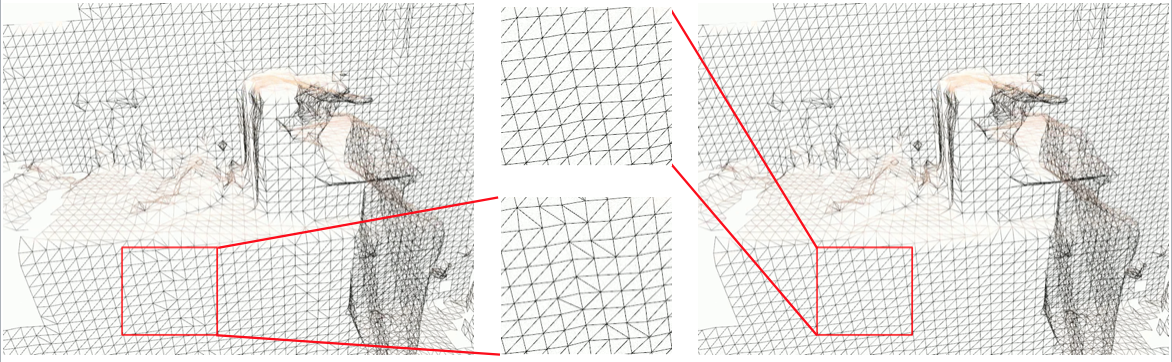}
			\label{fig:refinement-c}
		}
	\end{center}
	\caption{Mesh before (left) and after (right) refinement. {\it (a)}, simulated {\it office1} dataset. {\it (b)}, real world {\it copyroom} dataset. Besides areas zoomed in, similar refinements appear in the entire scene. The images are slightly degenerated due to compression.}\label{fig:refinement}
\end{figure}

\begin{table*}
	\begin{center}
		\begin{tabular}{|l|c||c|c|c|c||c|c|}
			\hline
			\multirow{3}{*}{Dataset} &
			\multirow{3}{*}{Frames} & 
			\multicolumn{4}{c||}{Time (ms)} &
			\multicolumn{2}{c|}{Memory (vertex count)} \\
			\cline{3-8}
			& & \multicolumn{2}{c|}{Ours} & \multicolumn{2}{c||}{Baseline} & \multirow{2}{*}{Ours} & \multirow{2}{*}{Baseline} \\
			\cline{3-6}
			& & Meshing & All & Meshing & All & & \\
			\hline\hline
			ICL/lv1 & 965 & \bf{6.06} & 7.20 (8.35) & 6.55 & 7.48 & {\bf85634} & 450903\\
			ICL/lv2 & 880 & \bf{6.89} & 7.85 (9.19) & 7.10 & 8.05 & {\bf105406} & 583986\\
			ICL/office1 & 965 & \bf{5.36} & 6.42 (7.71) & 5.88 & 6.78 & {\bf103574} & 577011\\	
			ICL/office2 & 880 & \bf{7.09} & 8.09 (9.39) & 7.70 & 8.54 & {\bf115885} & 619629\\
			TUM/household & 2486 & 10.94 & 12.21 (12.90) & \bf{9.61} & 10.83 & {\bf64198} & 327729 \\
			Zhou/copyroom & 5490 & {\bf 3.71} & 4.85 (5.68) & 3.95 & 4.94 & {\bf85699} & 446775\\
			Zhou/lounge & 3000 & {\bf4.03} & 5.05 (5.85) & 4.08 & 5.05 & {\bf62144} & 323562\\
			Zhou/burghers & 11230 & {\bf3.67} & 4.67 (5.48) & 3.76 & 4.72 & {\bf99976} & 532152\\
			\hline
		\end{tabular}
	\end{center}
	\caption{Average running time and total vertex consumption comparison of our method and baseline, at the resolution of 3cm. In implementation, our method requires an additional compressing operation before copying data to the rendering pipeline, while this step is ignored in the baseline due to our simplified implementation. Therefore running time (including all stages in \S\ref{sec:meshing} and rendering) of our method is displayed both without and with compressing stage, the latter in brackets.}\label{tab:table}
\end{table*}

\subsection{Qualitative Results}
Since the mesh generated by our method is in theory identical to the baseline method in geometric appearance, we do not focus on comparing mesh quality with baseline. Instead, we conduct experiments in two aspects, incremental reconstruction and refinement.

\textbf{Incremental reconstruction}. We process two sequences and render the global mesh against the newly modified mesh in sensor's frustum per frame. In addition, we visualize the existing duration of vertices, see Fig.\ref{fig:incremental}. In {\it household} (Fig.\ref{fig:incremental-a}) where sensor is generally far from the scene objects and motion blurs appear frequently, we accept a large scanning range and a coarse cube resolution to fuse in more valid data. When a loop closure emerges, most previous vertices are preserved, as shown in the color map. In {\it lounge} (Fig.\ref{fig:incremental-b}) where sensor are close to the objects and depth images are carefully captured, we run the program with a small scanning range and a high resolution. Most blocks are ignored during mesh generation, saving a large amount of time, while the mesh representing the whole scene remains consistent.

\textbf{Refinement.} Results with and without mesh refinement stage are compared both in the simulated dataset {\it office2} (Fig.\ref{fig:refinement-a}) with perfect sensor poses and depth images and the real-world dataset {\it copyroom} (Fig.\ref{fig:refinement-c}). The irregular `cracks' in the scenes are significantly reduced, leading to consistent triangle shapes and smooth planes. In the incremental reconstruction, the type of triangles sometimes suffer instability due to frequently flipped TSDF signs around, which could be further ameliorated by emphasizing temporal constraints.


\section{Conclusions and Future Work}
We propose a novel mesh representation with spatial hashed cube units that supports memory-efficient vertex sharing and time-efficient $O(1)$ accessing. Equipped with parallel algorithms, the data structure achieves considerable performance in the task of real-time scene reconstruction; additional refinement further improves the quality of mesh. 

There are several limitations in our pipeline apart from the memory and runtime issues discussed in \S\ref{sec:experiments}. First, we require precomputed accurate camera poses. When using online estimated pose from \eg \cite{murORB2}, inevitable drifts would cause the offset of TSDF value, leading to the shift of 3D models, reducing reconstruction quality. We also rely on the smoothing power of TSDF to eliminate noise from sensors, which is likely to filter out sharp details in scenes and might fail on very sparse depth data.

In the future, we plan to optimize the data structure and its manipulations. More sophisticated spatial hashing techniques might be used as proposed in \cite{KahlerPVM16}. We intend to open source the code as an useful tool for online reconstruction and mesh-based deformation and segmentation. In the research viewpoint, we are improving the data fusion stage considering the uncertainty from sensors and working on integrating localization module, utilizing the online generated mesh. A complete SLAM system would be our ultimate goal.

\end{document}